\documentclass{article}

\usepackage{microtype}
\usepackage{graphicx}
\usepackage{subcaption}
\usepackage{booktabs} 

\usepackage{hyperref}

\newcommand{\ours}{DLLG}
\usepackage[preprint]{icml2026}

\usepackage{amsmath}
\usepackage{amssymb}
\usepackage{mathtools}
\usepackage{amsthm}

\usepackage{adjustbox}
\usepackage[table]{xcolor}

\usepackage[capitalize,noabbrev]{cleveref}
\usepackage{enumitem}
\theoremstyle{plain}

\theoremstyle{definition}

\theoremstyle{remark}

\usepackage[textsize=tiny]{todonotes}

\icmltitlerunning{DLLG: Dynamic Logit-Level Gating of LLM Experts}

\begin{document}

\twocolumn[{%
  \icmltitle{DLLG: Dynamic Logit-Level Gating of LLM Experts}



  \icmlsetsymbol{internship}{$\dagger$}

  \begin{icmlauthorlist}
    \icmlauthor{Bingnan Li}{aws_agentic_ai,ucsd,internship}
    \icmlauthor{Zhaoyang Zhang}{aws_agentic_ai}
    \icmlauthor{Xiaoze Liu}{aws_agentic_ai,purdue,internship}
    \icmlauthor{Yantao Shen}{aws_agentic_ai}
    \icmlauthor{Shuli Jiang}{aws_agentic_ai}
    \icmlauthor{Shuo Yang}{aws_agentic_ai}
    \icmlauthor{Wei Xia}{aws_agentic_ai}
    \icmlauthor{Zhuowen Tu}{aws_agentic_ai}
    \icmlauthor{Stefano Soatto}{aws_agentic_ai}

  \end{icmlauthorlist}

  \icmlaffiliation{aws_agentic_ai}{AWS Agentic AI, Seattle, US}
  \icmlaffiliation{ucsd}{University of California San Diego, La Jolla, US}
  \icmlaffiliation{purdue}{Purdue University, West Lafayette, US}

  \icmlcorrespondingauthor{Zhaoyang Zhang}{ozhaozha@amazon.com}

  \icmlkeywords{Machine Learning, ICML}

  \vskip 0.3in
  \begin{center}
        \includegraphics[width=\textwidth]{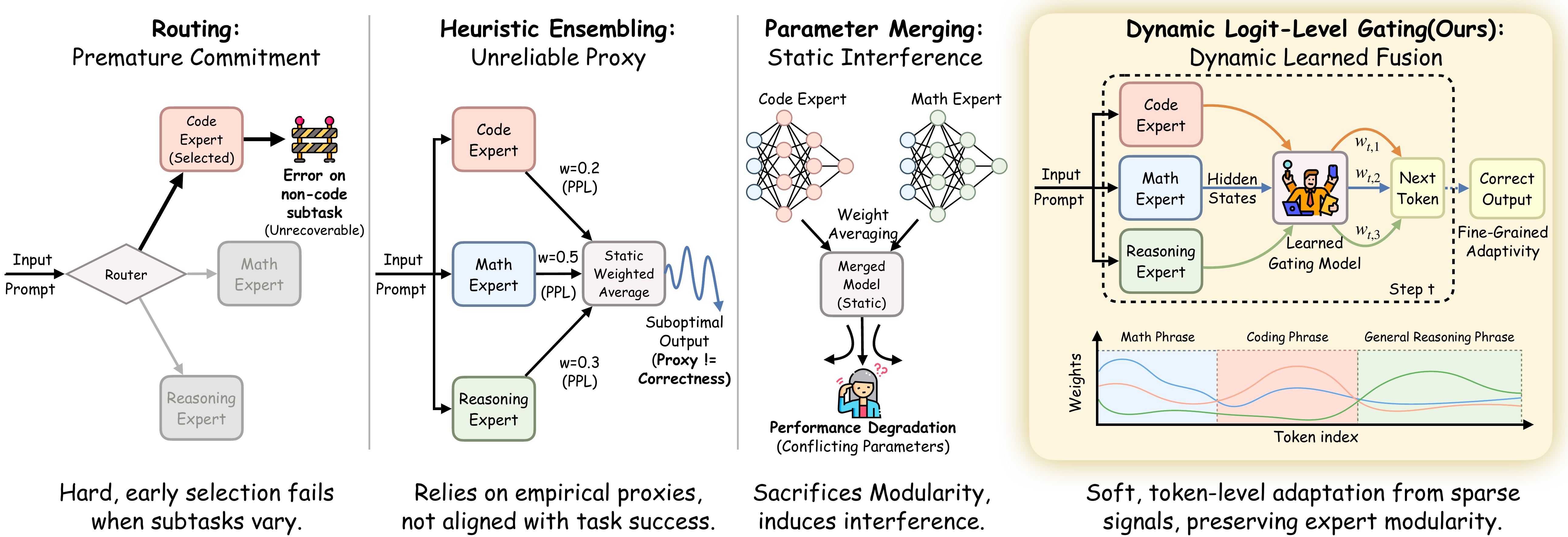}
        \captionof{figure}{
            Comparison of expert combination strategies.
            Routing relies on hard, early expert selection and fails when subtasks vary within a response.
            Heuristic ensembling uses inference-time proxy signals that are often misaligned with task correctness.
            Parameter merging statically fuses expert weights, sacrificing modularity and inducing interference.
            In contrast, \ours{} (ours) performs dynamic, token-level logit fusion using learned gating conditioned on expert hidden states, enabling fine-grained and recoverable expert utilization while preserving expert modularity.
        }
        \label{fig:teasor}
  \end{center}

}]

\printAffiliationsAndNotice{$\dagger$ The work was done during an internship at AWS Agentic AI. Bingnan Li is now with the University of California San Diego. Xiaoze Liu is now with the Purdue University.}

\begin{abstract}
Leveraging multiple specialized LLMs can combine complementary strengths, but existing approaches trade adaptability for stability: routing commits prematurely, heuristic ensembling depends on fragile proxies, and parameter merging introduces interference. 
We propose \ours{} (Dynamic Logit-Level Gating), a dynamic logit-level ensembling framework that learns token-level expert fusion from sparse response-level supervision.
A lightweight gating module predicts step-wise fusion weights, linking trajectory-level correctness to generation without token-level labels or expert retraining. 
Across diverse reasoning and code benchmarks, \ours{} consistently outperforms strong routing, heuristic ensembling, and parameter-merging baselines across model scales, highlighting learned logit-level fusion as a robust and scalable paradigm for integrating specialized experts.
\end{abstract}

\section{Introduction}

The landscape of Large Language Models (LLMs) has witnessed a proliferation of \emph{specialized models}, each optimized for narrow domains such as mathematical reasoning or code generation~\cite{qwen-coder,qwen-math}. While such specialists can outperform general-purpose models~\cite{gpt,llama} within their niches, relying on any single expert limits coverage across diverse tasks. This motivates a central question: \emph{how can we integrate independently trained specialists into a unified system that exploits their complementary strengths}~\cite{model-ensemble-survey,model-merging-survey}?

Prior work offers multiple directions, but existing approaches expose a persistent tension among \emph{adaptivity}, \emph{robustness}, and \emph{practicality}. \textbf{Routing-based methods} select one expert per input~\cite{FORC,EmbedLLM,Avengers,Bench-CoE,Hybird-LLM,MetaLLM,RTR,RouteLLM}, yet coarse early commitments are difficult to revise when expertise shifts within a single response; moreover, many require router training or calibration on benchmark data to ensure expert diversity~\cite{Bench-CoE,Avengers,RouterDC,RouteLLM}, an assumption rarely met in deployment. \textbf{Token-level ensembling} enables finer-grained combination~\cite{GaC,DeePEn,EVA,pack-of-llms,top-k-union,ABE}, from uniform averaging~\cite{GaC} to heuristics such as perplexity weighting~\cite{pack-of-llms} or top-$k$ token unions~\cite{top-k-union}, but these schemes remain largely empirical and rely on proxy signals that may not align with specialization or correctness. \textbf{Parameter-space merging} (e.g., model souping~\cite{model-souping} and task arithmetic~\cite{task-arithmetic}) yields a single model but sacrifices flexibility and often suffers destructive interference when combining disparate experts, making hyperparameter choices brittle~\cite{model-merging-survey}. Finally, while \textbf{Mixture-of-Experts} (MoE) architectures scale capacity via sparse gating~\cite{chen2023sparse,shazeer2017outrageously,dai2024deepseekmoe,jiang2024mixtral}, they typically require end-to-end joint training of the gate and experts, limiting their ability to leverage independently pre-trained, off-the-shelf specialists. Consequently, a key hurdle remains: effectively unifying these ``black-box'' experts into a single system without the prohibitive cost of expert retraining.

In this work, we introduce \textbf{\ours{}} (\textbf{D}ynamic \textbf{L}ogit-\textbf{L}evel \textbf{G}ating), a dynamic logit-level ensembling framework that learns fine-grained expert utilization from sparse supervision.
To avoid the premature commitment of routing, \ours{} performs autoregressive soft fusion: at each decoding step, a lightweight gate conditions on the prompt, the partial prefix, and trajectory-level hidden states from all experts to produce step-specific mixture weights for logit aggregation.
To replace brittle token-level heuristics, \ours{} learns the fusion rule via supervision. We train the gate with teacher forcing using response-level correctness labels from automatic verifiers (computed independently of the gate) and broadcast them to all tokens of the reference response.
\ours{} preserves expert modularity by freezing all expert parameters, avoiding interference and eliminating the need for test-time supervision or online rollouts.

Unlike MoE, which typically requires joint training of experts and gating, \ours{} treats specialists as plug-and-play frozen components and learns logit-level fusion from sparse supervision, achieving MoE-like adaptivity without expert retraining.
Our contributions are summarized as follows:

\begin{enumerate}[leftmargin=*,topsep=0pt,itemsep=0pt,parsep=0pt,partopsep=0pt]
    \item We propose \ours{}, a dynamic logit-level ensembling framework that bridges token-level generation and sparse response-level supervision. Using teacher-forced correctness labels, \ours{} learns fine-grained expert utilization without token-level annotation or online RL.
    
    \item We analyze the learned token-level fusion weights and show that the proposed gating mechanism dynamically adapts expert contribution during rollout, adjusting the contribution of different models in accordance with their specialization as the generation context evolves.
    
    \item We evaluate \ours{} on diverse reasoning and code benchmarks (GSM8K~\cite{gsm8k}, Minerva Math~\cite{minerva_math}, MATH~\cite{MATH}, Code-R1~\cite{code-r1}, HumanEval~\cite{humaneval}, MBPP~\cite{mbpp}, BBH~\cite{bbh}, BigCodeBench~\cite{bigcodebench}), showing consistent gains over routing, heuristic ensembling, and parameter-merging baselines across model scales.
\end{enumerate}

\section{Related Work}

We categorize existing strategies for combining specialized LLMs into four main paradigms: token-level aggregation, routing-based selection, parameter-space merging, and MoE.

\begin{figure*}[t]
    \centering
    \includegraphics[width=\linewidth]{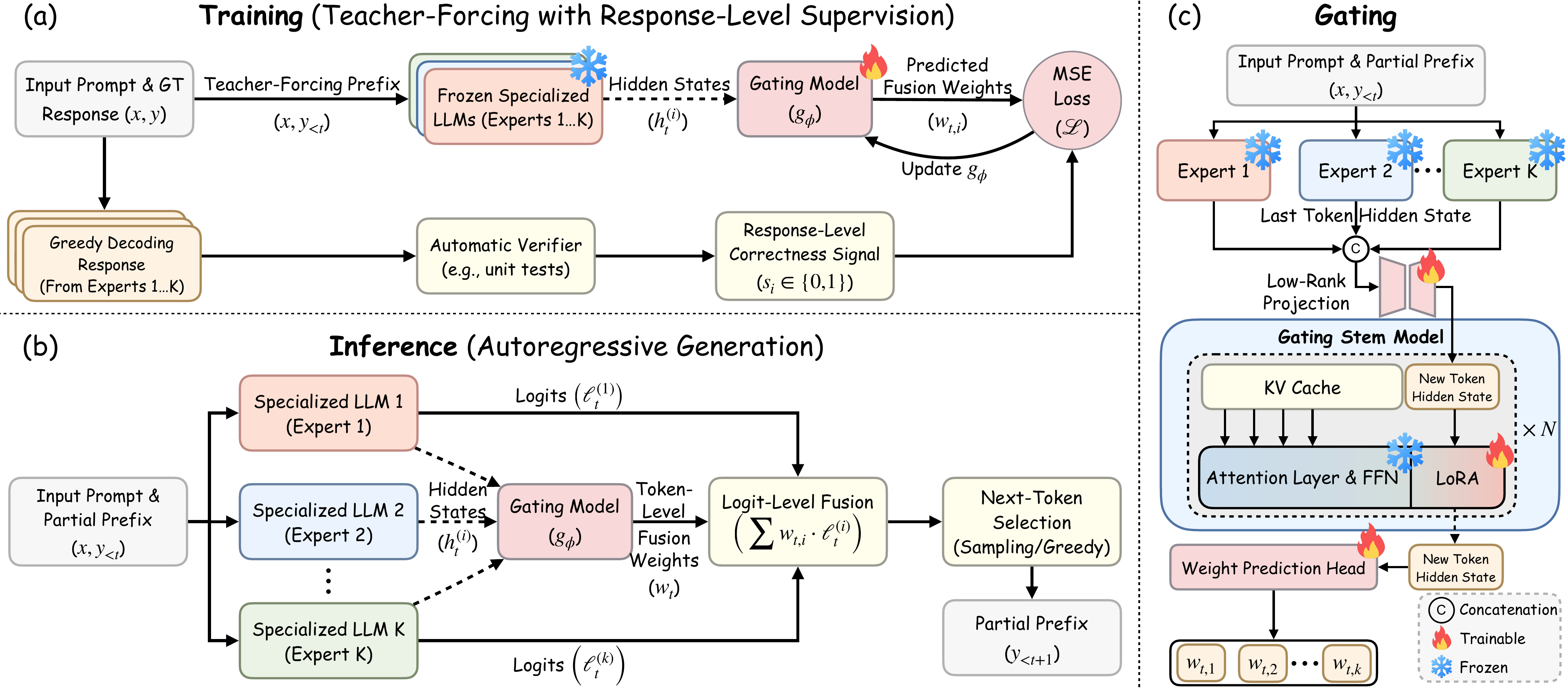}
    \caption{
Overview of \ours{}, illustrating training, inference, and the gating model architecture in a unified pipeline.
\textbf{(a) Training:} Frozen specialized LLMs are conditioned on ground-truth prefixes under teacher forcing to produce hidden states, which are fed into a lightweight gating model.
Response-level correctness signals, obtained from automatic verifiers applied to expert rollouts, supervise the gating model via an MSE objective.
\textbf{(b) Inference:} At each decoding step, the gating model predicts token-level fusion weights from expert hidden states, and expert logits are softly combined through logit-level fusion for autoregressive generation.
\textbf{(c) Gating model:} Hidden states from all experts are concatenated, projected into a shared embedding space, and processed by a gating stem with LoRA adapters and KV caching.
A weight prediction head outputs token-wise fusion weights, while all expert models remain frozen.
}
    \label{fig:pipeline}
    \vspace{-1em}
\end{figure*}

\begin{figure}[t]
    \centering
    \includegraphics[width=\linewidth]{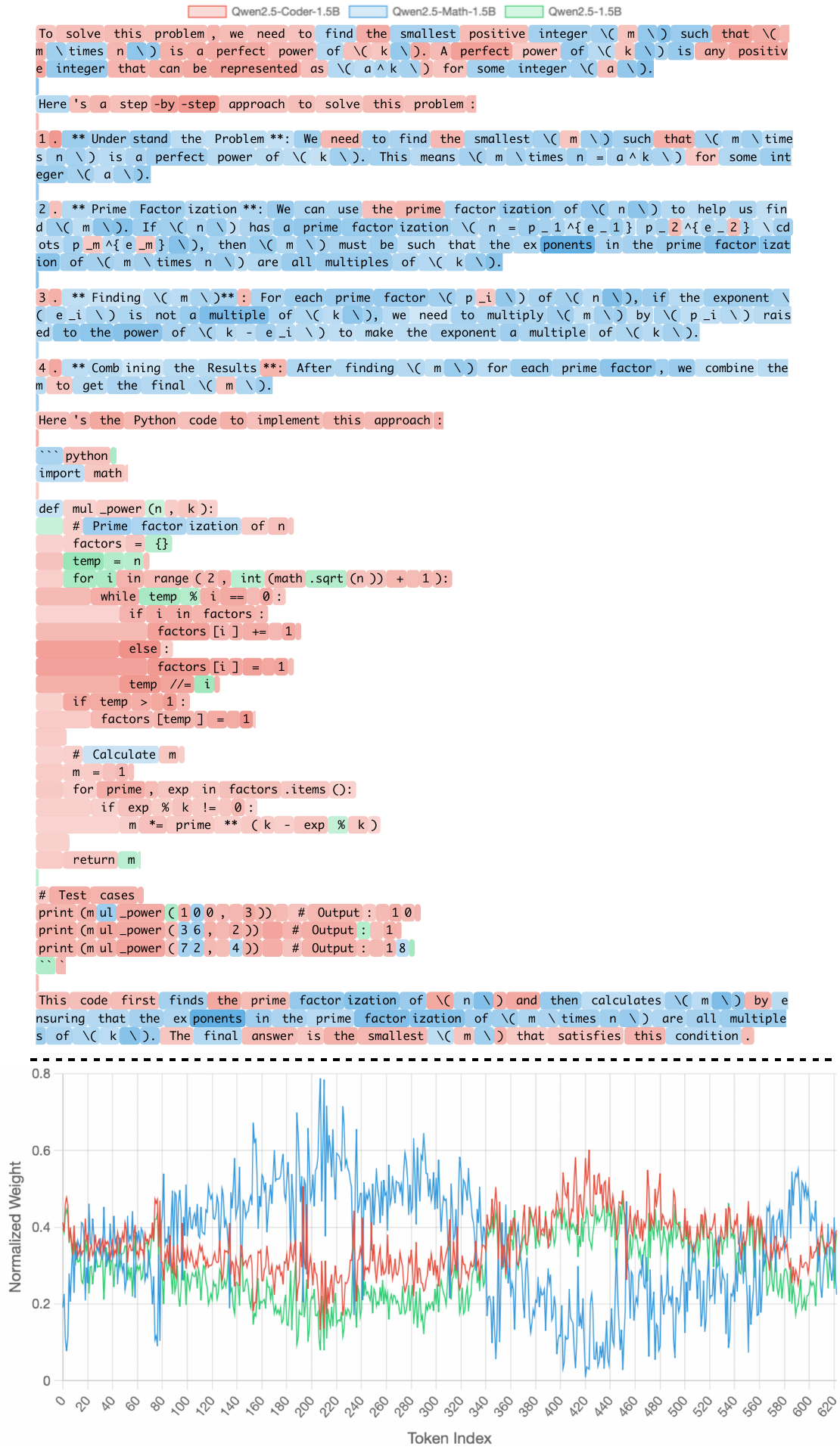}
\caption{
Token-level fusion behavior on a representative Code-R1 example, where the gating model dynamically adjusts expert fusion weights, with math-specialized experts dominating early reasoning stages and code-specialized experts becoming more prominent during code generation.
}
    \label{fig:token_fusion}
\vspace{-1em}

\end{figure}

\paragraph{Token-level aggregation} 

combines model outputs at the finest granularity and typically falls into two main categories: \emph{static} or \emph{heuristic}. \emph{Static approaches}, such as uniform averaging~\cite{GaC}, benefit from stability but lack the flexibility to account for varying expert specializations. \emph{Heuristic approaches} attempt to weight models dynamically using inference-time proxies, such as perplexity~\cite{pack-of-llms}, confidence scores~\cite{GaC}, or top-k token agreement~\cite{top-k-union}. However, these metrics are empirical proxies that often fail to align with true task correctness~\cite{ppl_not_reliable}. Additionally, while some works focus on \textbf{vocabulary alignment} for heterogeneous models~\cite{DeePEn,EVA,ABE}, they generally rely on simple fusion rules after projection. In contrast to heuristic or static ensembling approaches, our method learns token-level fusion weights directly from response-level supervision, providing a principled alternative that adapts expert contributions at a finer granularity while retaining the robustness advantages of logit-level ensembling.

\paragraph{Routing-based Expert Selection} 
aims to assign inputs to the most suitable model, typically operating at the coarse granularity of the entire prompt or response~\cite{RouterDC,EmbedLLM,Avengers,Bench-CoE,Hybird-LLM,MetaLLM,FORC,RTR}. Existing methods primarily rely on \emph{feature-based predictors} trained on prompt embeddings~\cite{RouterDC,EmbedLLM,Avengers} or \emph{statistical priors} derived from benchmark performance or preference signals~\cite{Bench-CoE,RouteLLM}. While some approaches focus on optimizing the performance-cost trade-off~\cite{Hybird-LLM,MetaLLM}, they generally suffer from two limitations. First, routing involves \emph{hard, premature commitment}: selecting a single expert early in generation leads to unrecoverable errors if the chosen model fails on a specific sub-task. Second, many routers require extensive \emph{benchmark-specific calibration} or test-set statistics to generalize~\cite{Bench-CoE,Avengers}, an assumption that rarely holds in realistic deployment. In contrast, \ours{} avoids hard selection entirely, utilizing soft, autoregressive fusion to adapt dynamically without relying on test-time ground truth.

\paragraph{Parameter-Space Merging} 
constructs a single model by fusing weights from multiple experts, typically assuming a shared architecture and initialization. Common techniques range from \emph{averaging-based methods} like model souping and SLERP~\cite{model-souping,slerp,Karcher-Mean} to \emph{arithmetic operations} on task vectors~\cite{task-arithmetic,ties,dare,della}. While computationally efficient at inference time, these approaches suffer from inherent \emph{static interference}: merging parameters optimized for distinct objectives often leads to performance degradation due to conflicting gradient directions~\cite{li2023deep,model-merging-survey}. Furthermore, the merged model is fixed after composition, sacrificing \emph{modularity} and the ability to dynamically leverage specific expert strengths during generation. By focusing on inference-time logit fusion, \ours{} bypasses parameter interference entirely, preserving the specialized capabilities of frozen experts.

\paragraph{Mixture-of-Experts (MoE)} is a foundational technique for increasing model capacity by activating only a subset of parameters per input through a router~\cite{chen2023sparse,shazeer2017outrageously,dai2024deepseekmoe,jiang2024mixtral}. While effective, standard MoE models rely on the simultaneous optimization of experts and routers to ensure balanced specialization~\cite{cai2025survey}. In contrast, \ours{} operates in a post-hoc ensembling regime where the experts are already fully specialized for distinct domains, such as math or code. By shifting the fusion to the logit level during inference, \ours{} inherits the dynamic flexibility of MoE’s gating logic but bypasses the need for expensive end-to-end training rollouts. This distinguishes \ours{} from existing routing methods that suffer from premature commitment, as our soft fusion allows for recoverable expert contributions at every decoding step.

\section{Method}
\label{sec:method}

In this section, we present \ours{} (Dynamic Logit-Level Gating), a dynamic logit-level ensembling framework for combining multiple specialized language models. \ours{} learns to assign token-level fusion weights under sparse response-level supervision, enabling fine-grained and adaptive expert contribution without requiring token-level annotations, hard expert selection, or expert retraining. The key idea is to train a lightweight gating model that predicts token-wise weights from expert trajectory representations, and to combine expert outputs through soft logit-level fusion at inference time. An overview of the training and inference pipeline of \ours{} is shown in \Cref{fig:pipeline}.

We first formalize the problem setting and the ensembling objective in~\Cref{sec:problem_formulation}. \Cref{sec:logit_ensemble_framework} introduces our logit-level fusion formulation, followed by the architecture of the gating model in~\Cref{sec:gating_model}. \Cref{sec:training_with_response-level_supervision} describes how the gating model is trained using response-level correctness signals under a teacher-forcing regime. Finally, \Cref{sec:inference_procedure} details the inference procedure, where experts are combined autoregressively using the learned token-level fusion weights.

\subsection{Problem Formulation}
\label{sec:problem_formulation}

We consider a setting with $K$ specialized large language models (experts)
$\{\mathcal{E}_i\}_{i=1}^{K}$ from the same model family, sharing a common tokenizer and vocabulary.
Let $x$ denote an input prompt, and let
$y = (y_1, \ldots, y_T)$ denote the corresponding output token sequence of length $T$,
where each $y_t \in \mathcal{V}$ and $\mathcal{V}$ is the shared vocabulary across all experts.
We use $y_{<t} = (y_1, \ldots, y_{t-1})$ to denote the prefix up to step $t-1$.

Given the input prompt $x$ and prefix $y_{<t}$, each expert $\mathcal{E}_i$ produces a hidden state
$\mathbf{h}_t^{(i)}$ and a next-token logit vector
$\boldsymbol{\ell}_t^{(i)} \in \mathbb{R}^{|\mathcal{V}|}$ autoregressively.
Our goal is to combine the outputs of these experts to generate a single output sequence that leverages their complementary strengths.

We assume access to response-level supervision for each expert, indicating whether the expert produces a correct response for a given input under an automatic verifier or a task-specific evaluation criterion.
This supervision is defined at the response level and does not provide token-level annotations.
The central challenge is therefore to learn fine-grained, token-level expert utilization under such sparse supervision.

\subsection{Logit-Level Ensembling Framework}
\label{sec:logit_ensemble_framework}

At decoding step $t$, each expert $\mathcal{E}_i$ produces a next-token logit vector
$\boldsymbol{\ell}_t^{(i)} \in \mathbb{R}^{|\mathcal{V}|}$ and a corresponding hidden state
$\mathbf{h}_t^{(i)}$, conditioned on the input prompt $x$ and the prefix $y_{<t}$.
We combine expert outputs at the logit level by computing a weighted sum:

\begin{equation}
\label{eq:logit_fusion}
\boldsymbol{\ell}_t = \sum_{i=1}^{K} w_{t,i} \, \boldsymbol{\ell}_t^{(i)},
\end{equation}

where $w_{t,i}$ denotes the token-level fusion weight assigned to expert $i$ at step $t$.
The combined logits $\boldsymbol{\ell}_t$ are then used for next-token prediction.

We do not require the fusion weights to sum to one.
Fusion weights are produced independently for each expert (e.g., via a sigmoid transformation) and may optionally be normalized without affecting the formulation.
This design enables soft and flexible expert contributions, in contrast to hard expert selection.

\subsection{Gating Model Architecture}
\label{sec:gating_model}

The core of \ours{} is a lightweight gating model that predicts token-level fusion weights from expert trajectory representations.
An overview of the gating architecture is shown in \Cref{fig:pipeline}(c).
At each decoding step $t$, the gating model processes hidden states produced by all experts and outputs step-specific fusion weights used for logit-level aggregation.

\paragraph{Expert representation aggregation.}
At decoding step $t$, each expert $\mathcal{E}_i$ produces a hidden state
$\mathbf{h}_{t}^{(i)} \in \mathbb{R}^{d}$.
We concatenate expert hidden states along the feature dimension and project them into a shared embedding space:
\begin{equation}
\mathbf{h}_{t}^{\text{cat}}
=
\text{Proj}\!\left(
\text{Concat}\left(\{\mathbf{h}_{t}^{(i)}\}_{i=1}^{K}\right)
\right),
\end{equation}
where $\text{Proj}(\cdot)$ denotes a learnable projection module.
In practice, this projection is implemented as a low-rank projection to reduce parameter count while preserving cross-expert information.

\paragraph{Gating backbone.}
The projected representation is processed by a gating stem model:
\begin{equation}
\mathbf{h}_{t}^{\text{gate}}
=
M_{\text{stem}}\!\left(\mathbf{h}_{t}^{\text{cat}}\right),
\end{equation}
which models trajectory-level context using an autoregressive architecture with cached key--value states.
We instantiate $M_{\text{stem}}$ using \textbf{Qwen2.5-0.5B-Instruct} and fine-tune it via LoRA adapters while keeping the backbone parameters frozen. We additionally provide an ablation study on different gating backbone architectures in \Cref{tab:ablation_gating_arch}.

\paragraph{Prediction of fusion weights.}
A lightweight prediction head maps the gating representation to unnormalized fusion scores, which are transformed into non-negative fusion weights via a sigmoid function:
\begin{equation}
\{w_{t,i}\}_{i=1}^{K}
=
\sigma\!\left(
\text{Head}\!\left(\mathbf{h}_{t}^{\text{gate}}\right)
\right),
\end{equation}
where $\sigma(\cdot)$ is applied element-wise.
The resulting fusion weights are used directly for logit-level ensembling without enforcing a simplex constraint.

\begin{table*}[t]
\centering
\setlength{\tabcolsep}{6pt}
\renewcommand{\arraystretch}{1.15}

\begin{adjustbox}{max width=\textwidth}
\begin{tabular}{lcccccccc}
\toprule
 & \multicolumn{4}{c}{\textbf{In-Domain}} & \multicolumn{3}{c}{\textbf{Out-of-Domain}} & \\
\cmidrule(lr){2-5} \cmidrule(lr){6-8}
\textbf{Method}
& GSM8K
& MinervaMath
& MATH
& Code-R1
& HumanEval
& MBPP
& BBH 
& Avg \\
\midrule

\multicolumn{9}{l}{\textbf{Experts}} \\
\midrule

\color{black!65}Qwen2.5-0.5B-Instruct
& \color{black!65}49.22 & \color{black!65}24.11 & \color{black!65}46.00 & \color{black!65}7.02 & \color{black!65}32.32 & \color{black!65}33.40 & \color{black!65}27.78 & \cellcolor{gray!18}\color{black!65}{31.41}\\

\color{black!65}Qwen2.5-Coder-0.5B-Instruct
& \color{black!65}31.25 & \color{black!65}11.16 & \color{black!65}22.00 & \color{black!65}7.02 & \color{black!65}54.88 & \color{black!65}35.80 & \color{black!65}27.20 & \cellcolor{gray!18}\color{black!65}{27.04} \\

\color{black!65}Dolphin3.0-Qwen2.5-0.5B
& \color{black!65}42.97 & \color{black!65}13.84 & \color{black!65}36.40 & \color{black!65}7.44 & \color{black!65}42.07 & \color{black!65}28.00 & \color{black!65}26.16 & \cellcolor{gray!18}\color{black!65}{28.13} \\
\midrule

\multicolumn{9}{l}{\textbf{Model Merging}} \\
\midrule

Linear~\cite{model-souping}
& 46.09 & 18.30 & 42.20 & 8.71 & 35.98 & 33.20 & 29.39 & \cellcolor{gray!18}{30.55} \\

SLERP
& 48.44 & 16.52 & 41.00 & 7.58 & 41.46 & 31.60 & 27.43 & \cellcolor{gray!18}{30.58} \\

Task Arithmetic~\cite{task-arithmetic}
& 26.56 & 5.80 & 14.20 & 5.48 & 43.90 & 33.00 & 29.51 & \cellcolor{gray!18}{22.64} \\

\midrule
\multicolumn{8}{l}{\textbf{Routing}} \\
\midrule
RouterDC~\cite{RouterDC}
& \underline{48.44} & 15.18 & 42.20 & 7.72 & 41.46 & 28.40 & 26.16 & \cellcolor{gray!18}{29.94} \\
EmbedLLM~\cite{EmbedLLM}
& 47.66 & \underline{19.20} & \textbf{44.60} & 8.71 & 32.31 & 33.80 & 27.78 & \cellcolor{gray!18}{30.58} \\
\midrule

\multicolumn{8}{l}{\textbf{Logits Ensemble}} \\
\midrule
GaC~\cite{GaC}
& \underline{50.78} & 16.07 & 42.40 & 8.85 & \underline{45.12} & \underline{35.40} & \underline{30.79} & \cellcolor{gray!18}{32.77} \\
Entropy Weighting
& 41.41 & 15.18 & 41.80 & 9.13 & \underline{45.12} & \textbf{36.00} & 27.66 & \cellcolor{gray!18}{30.90} \\
Pack of LLMs~\cite{pack-of-llms}
& 44.53 & 15.18 & 42.20 & 8.99 & \textbf{45.73} & 35.20 & 30.09 & \cellcolor{gray!18}{31.70} \\
Token Maj-Voting
& 45.31 & 16.07 & 43.00 & 6.60 & 43.30 & 33.20 & 27.20 & \cellcolor{gray!18}{30.67} \\
UniTe~\cite{top-k-union}
& \underline{50.78} & 15.63 & 42.60 & \underline{9.27} & \textbf{45.73} & \underline{35.40} & \textbf{30.90} & \underline{\cellcolor{gray!18}{32.90}} \\
\midrule
\textbf{Ours}
& \textbf{52.34} & \textbf{19.64} & \underline{43.40} & \textbf{9.55} & \textbf{45.73} & 35.00 & 30.56 & \textbf{\cellcolor{gray!18}{33.75}}\\
\bottomrule
\end{tabular}
\end{adjustbox}

\caption{Performance comparison across in-domain and out-of-domain benchmarks at the 0.5B scale (0-shot). Avg is the unweighted mean across all benchmarks. \textbf{Bold} and \underline{underline} denote the best and second-best results except experts, respectively.}
\label{tab:results_05b}
\vspace{-1em}
\end{table*}

\subsection{Training with Response-Level Supervision}
\label{sec:training_with_response-level_supervision}

Training the gating model is challenging due to the absence of token-level supervision.
We address this by leveraging response-level correctness signals as sparse supervision.

For each training example and expert $\mathcal{E}_i$, we obtain a binary correctness label
$s_i \in \{0,1\}$, indicating whether the expert's generated response is correct according to an automatic verifier or a task-specific evaluation criterion.
These response-level labels are computed independently of the gating model and are broadcast to all token positions of the corresponding ground-truth response during training.

We train the gating model in a teacher-forcing regime, where expert hidden states
$\mathbf{h}_t^{(i)}$ are computed conditioned on the ground-truth prefix $y_{<t}$.
At each token position $t$, the gating model predicts fusion weights $w_{t,i}$, which are supervised using the response-level signal $s_i$.
We minimize the following mean squared error objective:
\begin{equation}
\mathcal{L}
=
\frac{1}{T}
\sum_{t=1}^{T}
\sum_{i=1}^{K}
\big(w_{t,i} - s_i \big)^2,
\end{equation}
where $T$ denotes the length of the target output sequence.
The loss is averaged over all token positions and experts and it encourages the predicted token-level weights to align with the response-level correctness signal while allowing variation across tokens.

Teacher forcing decouples the learning of token-level fusion weights from the stochasticity of autoregressive generation, providing a low-variance and semantically aligned training signal under sparse supervision.

\subsection{Inference Procedure}
\label{sec:inference_procedure}

At inference time, \ours{} operates autoregressively.
Given the current prefix $y_{<t}$, all experts produce next-token logits and hidden states.
The gating model predicts step-specific fusion weights based on the expert representations, and the final logits are obtained via the logit-level combination in \Cref{sec:logit_ensemble_framework}.
The next token is sampled or selected from the combined logits, and the process repeats until termination.

This inference procedure enables soft and recoverable expert utilization at each decoding step, avoiding premature commitment to a single expert.
Since expert parameters remain unchanged, the method incurs minimal additional overhead and can be applied to arbitrary sets of specialized experts within a shared-tokenizer setting. When sufficient parallel GPU resources are available, the additional wall-clock latency over single-expert decoding can be limited, as experts are executed concurrently. 

\section{Experiments}
\begin{table*}[t]
\centering
\setlength{\tabcolsep}{4pt}
\renewcommand{\arraystretch}{1.15}

\begin{adjustbox}{max width=\textwidth}
\begin{tabular}{lccccccccc}
\toprule
 & \multicolumn{4}{c}{\textbf{In-Domain}} & \multicolumn{4}{c}{\textbf{Out-of-Domain}} & \\
\cmidrule(lr){2-5} \cmidrule(lr){6-9}
\textbf{Method} 
& GSM8K 
& MinervaMath
& MATH
& Code-R1 
& HumanEval
& MBPP
& BBH 
& BigCodeBench 
& Avg \\
\midrule
\multicolumn{10}{l}{\textbf{Experts}} \\
\midrule

\color{black!65}Qwen2.5-1.5B-Instruct 
& \color{black!65}67.97 & \color{black!65}30.36 & \color{black!65}69.40 & \color{black!65}15.03 & \color{black!65}53.66 & \color{black!65}49.80 & \color{black!65}38.19 & \color{black!65}3.40 & \color{black!65}\cellcolor{gray!18}40.98\\

\color{black!65}Qwen2.5-Math-1.5B-Instruct 
& \color{black!65}82.81 & \color{black!65}52.68 & \color{black!65}85.60 & \color{black!65}4.63 & \color{black!65}37.20 & \color{black!65}30.40 & \color{black!65}26.62 & \color{black!65}0.00 & \color{black!65}\cellcolor{gray!18}39.99\\

\color{black!65}Qwen2.5-Coder-1.5B-Instruct 
& \color{black!65}62.50 & \color{black!65}24.55 & \color{black!65}57.60 & \color{black!65}15.31 & \color{black!65}64.63 & \color{black!65}53.60 & \color{black!65}34.26 & \color{black!65}5.40 & \color{black!65}\cellcolor{gray!18}{39.73}\\

\midrule
\multicolumn{10}{l}{\textbf{Model Merging}} \\
\midrule

Linear~\cite{model-souping}
& 67.97 & 25.45 & 64.80 & 13.76 & 50.61 & 45.80 & 35.30 & 4.70 & \cellcolor{gray!18}38.55\\

SLERP
& 50.78 & 22.32 & 54.40 & 12.64 & 56.10 & 50.80 & 34.61 & 4.10 & \cellcolor{gray!18}35.72\\

Task Arithmetic~\cite{task-arithmetic}
& 50.00 & 11.61 & 35.00 & 12.36 & 46.95 & 42.00 & 31.02 & 3.40 & \cellcolor{gray!18}29.04\\

\midrule
\multicolumn{10}{l}{\textbf{Routing}} \\
\midrule

RouterDC~\cite{RouterDC}
& \textbf{82.81} & \underline{51.34} & 83.80 & 13.62 & \textbf{65.24} & \textbf{53.40} & 29.98 & \underline{6.10} & \cellcolor{gray!18}\underline{48.29}\\

EmbedLLM~\cite{EmbedLLM}
& 81.25 & \underline{51.34} & \textbf{84.60} & \underline{18.26} & 54.27 & 49.20 & 33.68 & \textbf{6.80} & \cellcolor{gray!18}47.43\\

\midrule
\multicolumn{10}{l}{\textbf{Logits Ensemble}} \\
\midrule

GaC~\cite{GaC}
& 81.25 & 45.09 & 80.20 & 16.71 & 60.98 & 50.60 & 34.84 & 2.70 & \cellcolor{gray!18}46.55\\

Entropy Weighting 
& 80.47 & 48.66 & 82.40 & \underline{18.26} & 62.19 & 51.60 & 33.10 & 4.10 & \cellcolor{gray!18}47.60\\

Pack of LLMs~\cite{pack-of-llms} 
& \underline{82.03} & 44.64 & 82.60 & 18.12 & \underline{64.02} & \underline{52.60} & 34.26 & 4.10 & \cellcolor{gray!18}47.80\\


Token Maj-Voting 
& 78.12 & 38.39 & 73.80 & 17.56 & 62.80 & 52.00 & \underline{35.42} & 3.40 & \cellcolor{gray!18}45.19\\

UniTe~\cite{top-k-union} 
& 80.47 & 45.98 & 80.20 & 18.12 & 61.59 & 50.00 & 35.53 & 1.40 & \cellcolor{gray!18}46.66\\

\midrule
\textbf{Ours} 
& \textbf{82.81} & \textbf{51.79} & \underline{84.40} & \textbf{19.96} & \textbf{65.24} & \underline{52.60} & \textbf{36.11} & 4.10 & \cellcolor{gray!18}\textbf{49.63}\\
\bottomrule
\end{tabular}
\end{adjustbox}

\caption{Performance comparison across in-domain and out-of-domain benchmarks at the 1.5B scale (0-shot). Avg is the unweighted mean across all benchmarks. \textbf{Bold} and \underline{underline} denote the best and second-best results except experts, respectively.}
\label{tab:main_results}
\vspace{-1em}
\end{table*}

\subsection{Implementation Details}
We evaluate \ours{} using multiple specialized language models from the Qwen2.5 family~\cite{qwen,qwen-coder,qwen-math}, ensuring a shared tokenizer and aligned vocabularies. The gating model is built on top of the Qwen2.5-0.5B-Instruct model and trained with LoRA~\cite{lora}. During training, we adopt teacher forcing and optimize the gating model using a mean squared error objective. At inference time, all experts run in parallel and  are combined autoregressively through soft logit-level fusion as described in~\Cref{sec:method}. All experiments are conducted on 8 NVIDIA RTX H200 GPUs. For more details, please refer to Appendix~\Cref{app:implementation_details}.

\subsubsection{Training Data}
We collect training data from GSM8K~\cite{gsm8k}, MATH~\cite{MATH} and Code-R1~\cite{code-r1} training set, resulting in approximately 24K training examples covering mathematical reasoning and code generation. The response-level supervision signals are obtained using task-specific evaluation criteria or automatic verifiers, depending on the benchmark.

\begin{table*}[t]
\centering
\setlength{\tabcolsep}{6pt}
\renewcommand{\arraystretch}{1.15}

\begin{adjustbox}{max width=\textwidth}
\begin{tabular}{lccccccccc}
\toprule
 & \multicolumn{4}{c}{\textbf{In-Domain}} & \multicolumn{4}{c}{\textbf{Out-of-Domain}} & \\
\cmidrule(lr){2-5} \cmidrule(lr){6-9}
\textbf{Gating Model Design}
& GSM8K
& MinervaMath
& MATH
& Code-R1
& HumanEval
& MBPP
& BBH
& BigCodeBench 
& Avg \\
\midrule

MLP Head
& \textbf{82.81} & 47.32 & 81.60 & 17.42 & \textbf{65.85} & 52.00 & 32.87 & 5.40 & \cellcolor{gray!18}48.16 \\

Cross Attn Head
& 81.25 & 50.45 & 83.40 & 17.14 & 65.24 & 52.80 & 33.45 & 4.10 & \cellcolor{gray!18}48.48 \\

\shortstack[l]{Aux-LLM+MLP}
& \textbf{82.81} & 47.77 & 82.00 & \textbf{21.21} & 64.63 & \textbf{53.60} & 33.80 & \textbf{6.10} & \cellcolor{gray!18}48.99\\

\shortstack[l]{LowRank+Aux-LLM+MLP}
& \textbf{82.81} & \textbf{51.79} & \textbf{84.40} & 19.96 & 65.24 & 52.60 & \textbf{36.11} & 4.10 & \cellcolor{gray!18}\textbf{49.63}\\

\bottomrule
\end{tabular}
\end{adjustbox}

\caption{Ablation on gating model architecture under the 1.5B scale setting.
Aux-LLM denotes a lightweight auxiliary language model (Qwen-0.5B-Instruct), and LowRank denotes an additional low-rank projection applied to expert representations.}
\label{tab:ablation_gating_arch}
\vspace{-1em}
\end{table*}

\subsubsection{Benchmarks and Baselines}

We evaluate our method on a diverse set of reasoning and code generation benchmarks, covering both in-domain and out-of-domain settings. In-domain benchmarks include GSM8K~\cite{gsm8k}, MinervaMath~\cite{minerva_math}, MATH~\cite{MATH}, and Code-R1~\cite{code-r1}. Out-of-domain benchmarks include HumanEval~\cite{humaneval}, MBPP~\cite{mbpp}, BBH~\cite{bbh}, and BigCodeBench~\cite{bigcodebench}. For fair comparison, we utilized lm-evaluation-harness and official evaluation kits of Qwen2.5-Math, Code-R1 and BigCodeBench.

We compare \ours{} against a broad range of baselines, including individual parameter-space model merging methods (Linear~\cite{model-souping}, SLERP~\cite{slerp}, Task Arithmetic~\cite{task-arithmetic}), routing-based approaches (RouterDC~\cite{RouterDC}, EmbedLLM~\cite{EmbedLLM}), and logit-level ensembling methods such as uniform averaging (GaC~\cite{GaC}), entropy-based weighting, Pack of LLMs~\cite{pack-of-llms}, token majority voting, and UniTe~\cite{top-k-union}. All baselines are evaluated under the same 0-shot setting for fair comparison.

\subsection{Multi-Scale Experiments}

\subsubsection{0.5B Experiments}

\Cref{tab:results_05b} reports results at the smaller 0.5B scale, which serves as a challenging low-capacity setting where individual experts are relatively weak. Despite this constraint, \ours{} achieves the best average performance among all baselines, demonstrating that the benefits of learned token-level fusion extend beyond high-capacity regimes.

Compared to routing-based and heuristic ensembling methods, \ours{} exhibits more robust behavior in this low-capacity setting. When experts are error-prone or unevenly specialized, brittle routing decisions and unreliable proxy-based weighting strategies can severely degrade performance. In contrast, \ours{} learns to allocate expert contributions dynamically at the token level, mitigating early mistakes and enabling recoverable expert utilization throughout generation.

Overall, these results indicate that \ours{} remains effective even when individual experts are weak, highlighting its robustness in low-capacity regimes and motivating evaluation at larger model scales.

\subsubsection{1.5B Experiments}

\Cref{tab:main_results} summarizes the results at the 1.5B scale across both in-domain and out-of-domain benchmarks. At this larger scale, \ours{} consistently achieves the best average performance and ranks among the top methods on nearly all individual tasks, confirming that the advantages observed at smaller scales persist as model capacity increases.

Compared to routing-based methods, \ours{} demonstrates substantially more stable performance across diverse benchmarks. While routing approaches such as RouterDC~\cite{RouterDC} and EmbedLLM~\cite{EmbedLLM} perform competitively on selected in-domain tasks, their performance degrades when expert strengths vary within a single response, as in mixed reasoning or code-generation benchmarks. This limitation stems from hard or coarse-grained expert selection: once an expert is chosen early in the generation process, errors introduced by suboptimal routing decisions are difficult to recover. In contrast, \ours{} performs soft, token-level fusion throughout decoding, allowing expert contributions to adapt dynamically as the generation context evolves.

Relative to logit-level ensembling baselines, \ours{} consistently outperforms uniform averaging and heuristic weighting strategies~\cite{pack-of-llms,top-k-union}, including entropy-based weighting and perplexity-driven methods. These approaches rely on empirical proxies that are only loosely correlated with task correctness and fail to capture fine-grained specialization at the token level. By contrast, \ours{} learns fusion weights directly from response-level correctness supervision, enabling more effective exploitation of expert complementarity across both reasoning-intensive tasks (e.g., GSM8K~\cite{gsm8k}, MinervaMath~\cite{minerva_math}, MATH~\cite{MATH}) and code-generation benchmarks (e.g., Code-R1~\cite{code-r1}, HumanEval~\cite{humaneval}).

Finally, parameter-space model merging methods~\cite{model-souping,slerp,task-arithmetic} underperform inference-time ensembling approaches by a substantial margin. This gap highlights the challenges of static interference when merging experts with heterogeneous specializations. Since merged models are fixed after composition, they lack the flexibility to adapt expert contributions during generation, whereas \ours{} preserves expert modularity and dynamically integrates expert outputs at inference time.
\subsection{Ablation Study}

To better understand the design choices of \ours{}, we conduct an ablation study on the gating model architecture under the 1.5B scale setting, with results summarized in~\Cref{tab:ablation_gating_arch}.

We first observe that simple gating designs, such as a standalone MLP head, fail to achieve competitive performance. This indicates that naively predicting fusion weights without sufficiently expressive expert representations is insufficient for effective token-level fusion. Introducing a cross-attention-based gating head improves performance, confirming the importance of modeling interactions between experts.

Further gains are achieved by incorporating a lightweight auxiliary language model (Aux-LLM) into the gating mechanism. Conditioning the gating model on richer trajectory-level representations enables more accurate estimation of expert utility, particularly on reasoning-heavy benchmarks such as MinervaMath~\cite{minerva_math} and MATH~\cite{MATH}. Adding a low-rank projection layer further stabilizes training and improves out-of-domain generalization, suggesting that controlled dimensionality reduction helps mitigate overfitting to specific expert behaviors.

Overall, the full \ours{} configuration achieves the strongest and most balanced performance across both in-domain and out-of-domain benchmarks. These results confirm that effective token-level fusion requires not only response-level supervision, but also a sufficiently expressive yet lightweight gating architecture capable of modeling expert trajectories without introducing excessive overhead.

\subsection{Analysis of Token-Level Fusion}
\label{sec:token_level_analysis}

To better understand how \ours{} utilizes different experts during generation, we analyze the learned token-level fusion behavior through qualitative visualization.
\Cref{fig:token_fusion} presents a representative example from the Code-R1~\cite{code-r1} benchmark, illustrating how expert contributions evolve across decoding steps.

As shown in the top panel of \Cref{fig:token_fusion}, different experts dominate the fused logits at different token positions.
In the early stages of generation, when the response primarily involves mathematical reasoning or symbolic manipulation, the math-specialized expert tends to receive higher fusion weights.
As the generation progresses and transitions into code synthesis and implementation details, the gating model gradually shifts emphasis toward the code-specialized expert.
This behavior highlights the limitation of coarse-grained routing~\cite{RouterDC,EmbedLLM} or static ensembling~\cite{pack-of-llms,top-k-union}, which cannot revise early decisions when subtask requirements change within a single response.

The bottom panel visualizes the normalized fusion weights predicted by the gating model over decoding steps.
Rather than remaining static or collapsing to a single expert, the weights vary smoothly and adapt to the evolving generation context.
This demonstrates that \ours{} learns fine-grained, context-aware expert fusion aligned with subtask structure, enabling recoverable expert utilization throughout generation.

Additional qualitative examples exhibiting similar fusion patterns across different inputs are provided in Appendix~\Cref{app:more_vis}.

\section{Conclusion}

In this work, we introduced \ours{}, a dynamic logit-level ensembling framework that learns token-level fusion weights from sparse response-level supervision. By training a lightweight gating model under a teacher-forcing regime and combining expert outputs through soft logit-level fusion, \ours{} enables fine-grained and recoverable expert utilization without requiring token-level annotations, hard expert selection, or expert retraining.

Across a broad range of reasoning and code generation benchmarks and multiple model scales, \ours{} consistently outperforms strong routing, heuristic ensembling, and parameter-space merging baselines, demonstrating robust improvements across both in-domain and out-of-domain settings. Our analysis further shows that the learned fusion weights evolve dynamically during rollout, enabling effective and recoverable expert utilization within a single response.

\ours{} focuses on the fusion mechanism itself and assumes aligned token-level representations among experts. Extending the framework to settings with heterogeneous tokenizers and integrating vocabulary alignment techniques is a natural direction for future work. More broadly, we believe that learning fine-grained expert utilization from weak supervision offers a promising and practical paradigm for leveraging increasingly specialized language models.

\section*{Impact Statement}

This paper presents a method for improving the robustness and adaptability of large language models through learned logit-level ensembling. The primary goal of this work is to advance the field of machine learning by enabling more effective utilization of specialized models without additional supervision or retraining.

As with many techniques that improve the performance and flexibility of language models, the proposed approach may contribute to downstream applications that rely on such models. While these applications could have broad societal impacts depending on their specific use cases, we do not identify any unique or direct ethical concerns introduced by the method itself beyond those commonly associated with large language models. We therefore believe that the societal implications of this work are consistent with those of existing research in model ensembling and language model deployment.


\bibliography{icml}
\bibliographystyle{icml2026}

\newpage
\appendix
\onecolumn
\section{More Implementation Details}
\label{app:implementation_details}

\subsection{Training}
The gating model is trained using the AdamW optimizer with a learning rate of $1\times10^{-3}$.
Training is performed for a total of 1{,}000 optimization steps.
All LoRA adapters are configured with rank 16. The projection rank is set to be 64.
The batch size is fixed to 32 across all experiments.
Unless otherwise specified, all remaining optimization hyperparameters follow standard settings.

\subsection{Evaluation}
All evaluations are conducted under a strictly zero-shot setting. 

Due to the computational cost of multi-model logit-level ensemble inference, we evaluate several large benchmarks on fixed subsets. Unless otherwise specified, subsets are selected by taking the first $N$ examples from the original benchmark order, without any filtering based on model outputs.

For benchmarks evaluated using \texttt{lm-evaluation-harness}, we report results on \texttt{gsm8k\_cot} (first 128 examples), \texttt{minerva\_math} (first 32 examples for each sub-task), \texttt{humaneval\_instruct} (full set), \texttt{mbpp\_instruct} (full set), and \texttt{bbh\_zeroshot} (first 32 examples for each sub-task). The MATH benchmark is evaluated using the official Qwen2.5-Math evaluation toolkit on the first 500 examples. For BigCodeBench, we evaluate on both the \texttt{complete} split and the \texttt{hard} subset.

For Code-R1, we evaluate all 712 examples using the official evaluation framework with a minor modification to the reward function. Specifically, the weights of all format-related reward components are set to zero, and correctness is determined solely based on the extracted code and its execution results. This setting isolates functional correctness from formatting-related factors.

Across all benchmarks, the maximum generation length is fixed to 3{,}072 tokens.
All evaluations apply the default chat templates provided by the corresponding evaluation toolkit for each task.

\section{More Fusion Weights Visualizations}
\label{app:more_vis}
To further illustrate the behavior of the learned token-level fusion in \ours{}, we present additional qualitative visualizations of fusion weights across different problem instances.
These examples complement the analysis in \Cref{sec:token_level_analysis} and demonstrate that the observed fusion patterns are consistent across diverse inputs.

\begin{figure}[t]
    \centering
    \includegraphics[width=1\linewidth]{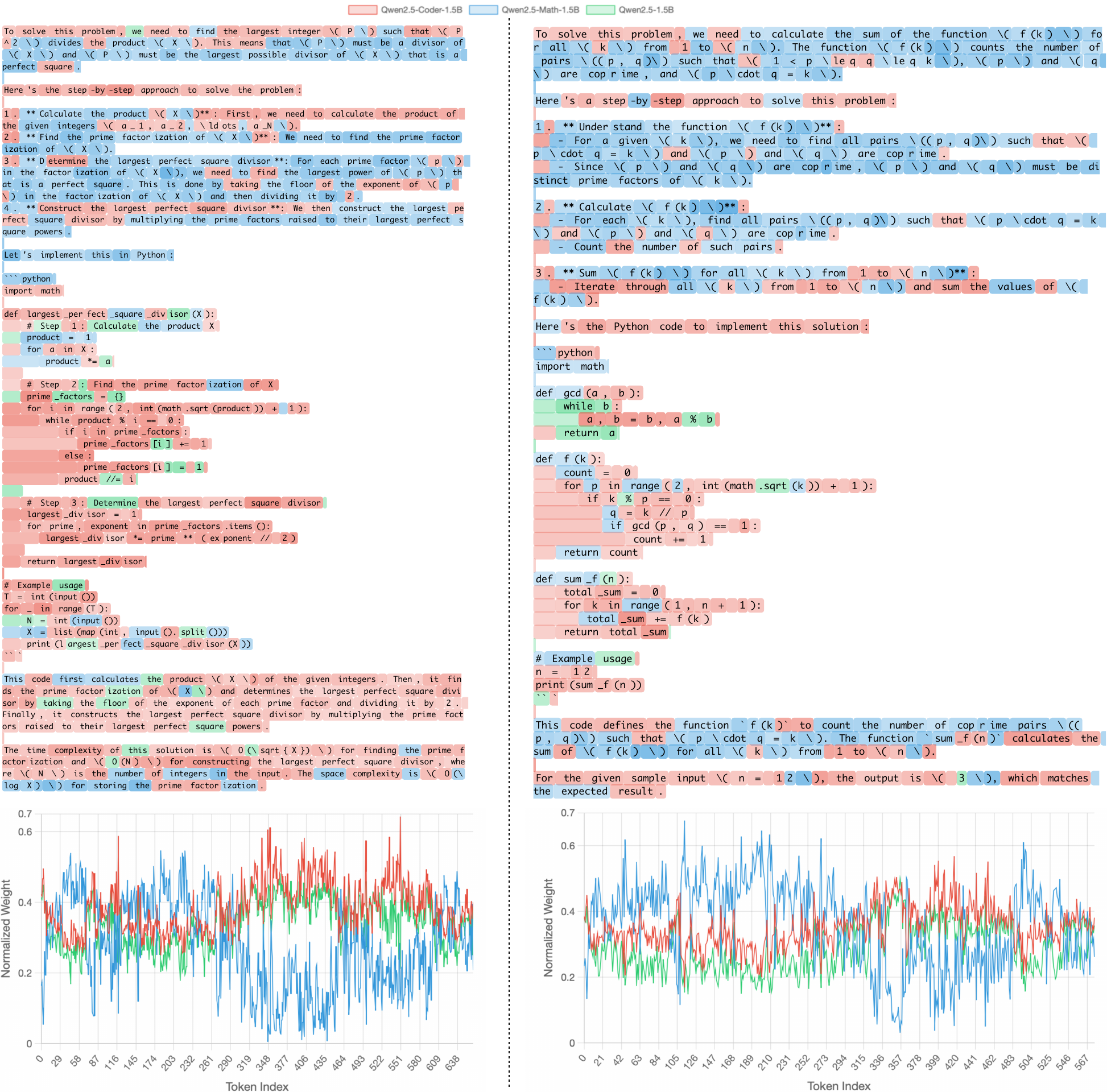}
\caption{
Additional visualizations of token-level weights in \ours{} across different examples.
For each example, the top panel highlights token-wise expert dominance in the fused output, and the bottom panel shows the normalized fusion weights over decoding steps.
Across examples, math-specialized experts tend to dominate during early reasoning-heavy segments, while code-specialized experts become more prominent during later code-generation phases.
The smooth transitions between experts illustrate \ours{}’s ability to perform context-aware, token-level fusion rather than static weighting or hard expert selection.
}
    \label{fig:more_weight_vis}
\end{figure}

Figure~\ref{fig:more_weight_vis} shows representative examples from reasoning- and code-centric tasks.
For each example, the top panel visualizes token-level expert dominance in the fused output, while the bottom panel plots the normalized fusion weights predicted by the gating model over decoding steps.

A clear and recurring pattern emerges across examples.
In the early stages of generation, when the response primarily involves mathematical reasoning or symbolic manipulation, the math-specialized expert receives higher fusion weights and dominates the combined logits.
As the generation transitions into code synthesis, implementation details, or executable logic, the gating model progressively shifts emphasis toward the code-specialized expert.
This transition occurs smoothly over tokens rather than through abrupt switches, reflecting fine-grained and recoverable expert utilization.

Importantly, the fusion weights do not collapse to static values or a single dominant expert.
Instead, \ours{} dynamically reallocates expert influence in a context-dependent manner, enabling different experts to contribute where they are most effective within a single response.
These qualitative results further support that the learned gating mechanism captures subtask structure and generalizes beyond individual examples, consistently performing token-level fusion aligned with the evolving generation context.

\section{Limitations}
\subsection{Latency and resource cost}
A practical limitation of DLLG is its increased inference-time resource cost. Unlike standard single-model decoding, DLLG requires all experts to be executed at every decoding step before the next token can be selected. In our implementation, experts are executed in parallel across multiple GPUs, which helps limit the additional wall-clock latency compared with single-expert decoding. However, this comes at the cost of higher aggregate computation, memory usage, and hardware requirements. If sufficient parallel resources are not available, executing experts sequentially would directly increase per-token latency. 

These resource costs define the scope of our empirical comparisons. Our goal is to study whether specialized models can be dynamically combined at the token level, rather than to establish compute-optimality against a single larger model under a matched inference budget. We therefore do not claim that DLLG is preferable to scaling a single model under the same total compute or memory budget. Compute-matched comparisons with larger single models are left to future work.

\subsection{Serving limitations}
DLLG is currently implemented on top of HuggingFace Transformers, which provides explicit control over the rollout loop and simplifies KV-cache management. This is important because DLLG requires all experts to advance in lock-step: at each decoding step, it collects expert logits and hidden states, predicts token-level fusion weights, merges logits before sampling, and feeds the same generated token back to all experts.

This design is not directly compatible with off-the-shelf \texttt{vLLM} serving. While \texttt{vLLM} is highly optimized for standard single-model inference, its rollout loop, scheduling, sampling, and KV-cache management are largely encapsulated inside the engine. DLLG, in contrast, requires a cross-model synchronization point before every sampling step and aligned KV caches across all experts. Running experts as separate \texttt{vLLM} engines would further introduce token-level communication overhead and complicate cache alignment under continuous batching. We therefore leave optimized \texttt{vLLM}-style serving support for DLLG to future work.

\end{document}